# Real Time System for Facial Analysis

Janne Tommola, Pedram Ghazi, Bishwo Adhikari, and Heikki Huttunen
Tampere University of Technology, Finland

## I. INTRODUCTION

Most signal or image processing algorithms should be designed with real-time execution in mind. Most use cases compute on an embedded platform while receiving streaming data as a constant data flow. In machine learning, however, the real time deployment and streaming data processing are less often a design criterion. Instead, the bulk of machine learning is executed offline on the cloud without any real time restrictions. However, the real time use is rapidly becoming more important as deep learning systems are appearing into, for example, autonomous vehicles and working machines.

In this work, we describe the functionality of our demo system integrating a number of common real time machine learning systems together. The demo system consists of a screen, webcam and a computer, and it estimates the age, gender and facial expression of all faces seen by the webcam. A picture of the system in use is shown in Figure 1. There is also a Youtube video at https://youtu.be/Kfe5hKNwrCU and the code is freely available at https://github.com/mahehu/TUT-live-age-estimator.

Apart from serving as an illustrative example of modern human level machine learning for the general public, the system also highlights several aspects that are common in real time machine learning systems. First, the subtasks needed to achieve the three recognition results represent a wide variety of machine learning problems: (1) *object detection* is used to find the faces, (2) age estimation represents a *regression problem* with a real-valued target output (3) gender prediction is a *binary classification* problem, and (4) facial expression prediction is a *multi-class classification* problem. Moreover, all these tasks should operate in unison, such that each task will receive enough resources from a limited pool.

In the remainder of this paper, we first describe the system level multithreaded architecture for real time processing in Section II. This is followed by detailed discussion individual components of the system in Section III. Next, we report experimental results on the accuracy of each individual recognition component in Section IV, and finally, discuss the benefits of demonstrating the potential of modern machine learning to both general public and experts in the field.

## II. SYSTEM LEVEL FUNCTIONALITY

The challenge in real-time operation is that there are numerous components in the system, and each uses different amount of execution time. The system should be designed such that the operation appears smooth, which means that the most visible tasks should be fast and have the priority in scheduling.

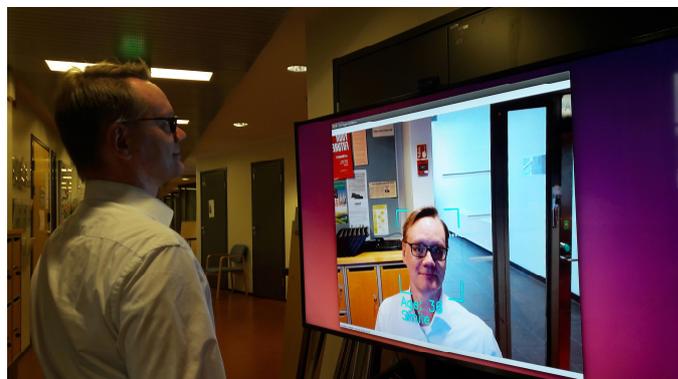

**Figure 1.** Demo system recognizes the age, gender and facial expression in real time.

The system is running in threads, as illustrated in Figure 2. The whole system is controlled by the upper level *controller and visualization thread,* which owns and starts the sub-threads dedicated for individual tasks. The main thread holds all data and executes the visualization loop showing the recognition results to the user at 25 frames per second.

The recognition process starts from the *grabber thread*, which is connected to a webcam. The thread requests video frames from camera for feeding them into a FIFO buffer located inside the controller thread. At grab time, each frame is wrapped inside a class object, which holds the necessary meta data related to each frame. More specifically, each frame is linked with a timestamp and a flag indicating whether the face detection has already been executed and the locations (bounding boxes) of all found faces in the scene.

The actual face analysis consists of two parts: face detection and face analysis. The detection is executed in the *detection thread*, which operates asynchronously, requesting new non-processed frames from the controller thread. After face detection, the locations of found faces are sent to the controller thread, which then matches each new face with all face objects from the previous frames using straightforward centroid tracking. Tracking allows us to average the estimates for each face over a number of recent frames.

The detection thread operates on the average faster than the frame rate, but sometimes there are delays due to high load on the other threads. Therefore, the controller thread holds a buffer of the most recent frames, in order to increase the flexibility of processing time.

The *recognition thread* is responsible for assessing the age, gender and facial expression of each face crop found from the image. The thread operates also in an asynchronous mode, requesting new non-processed (but face-detected) frames from the controller thread. When a new frame is received, the thread first aligns the face with a face template. This is done by detecting 68 facial landmarks using an ensemble of regression trees [1] as implemented in the *dlib* computer vision library (http://dlib.net/). After face alignment, we pass each aligned face to three separate networks: age recognizer (Section III-B), gender recognizer (Section III-C), and expression recognizer (Section III-D). Typically, each network takes approximately 200 ms for the forward pass, so in total the thread consumes about 600 ms per cropped face. Naturally, the amount to time grows linearly with the number of faces in the scene. However, the system still appears fast and responsive, because the camera view and all detections are real-time. Only the recognition results are updated only about once per second. The recognition thread is also optimized to evaluate age and gender less frequently than the expression. This is because age and gender are supposed to be constant, and users expect a quick response to their expressions. This reduces the average time per face by about 50% to approximately 300 ms.

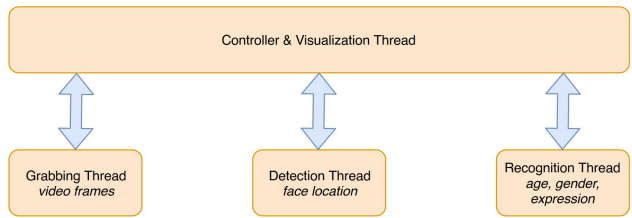

**Figure 2.** System architecture

The system is implemented in Python, and uses OpenCV, Tensorflow and dlib libraries for all computation. More specifically, the face detector is trained with Tensorflow Object Detection API and used through OpenCV, face alignment uses dlib, and all recognition networks run on Tensorflow via Keras front end. The system runs on an i7 2600 CPU only: In our experiments we discovered that the GPU does not significantly speed up the computation, but degrades the frame rate of visualization as GPU rendering is slowed down.

### III. COMPONENTS OF THE SYSTEM

#### A. Face detection

The face detection uses the SSD detector [2] with MobileNet [3] backbone, depth parameter $\alpha = 0.75$ and input size 240x180. The depth parameter was chosen for fast performance and the input size matches the camera's aspect ratio 4:3. The network is initialized using COCO-pretrained weights and trained with FDDB [4] face database.

#### B. Age recognition

The age estimation uses MobileNet with depth parameter $\alpha = 1.0$. The network is initialized using Imagenet-pretrained weights and fine-tuned in two stages: first with the large but noisy 500K IMDB-WIKI dataset [5] and then using the small but accurate CVPR2016 LAP challenge dataset [6].

#### C. Gender recognition

MobileNet with $\alpha = 1.0$ is used again. The last three layers were removed and replaced by a layer containing a single neuron and the network is trained from scratch in two stages: first with the 500K IMDB-WIKI dataset [5] and then fine-tuned with the CVPR2016 LAP challenge dataset [6].

#### D. Facial expression recognition

MobileNet with $\alpha = 1.0$ is used here, as well. The network is initialized with Imagenet pre-trained weights and fine-tuned with AffectNet database [7] containing 7 different expressions.

### IV. EXPERIMENTS

The accuracies of each component are shown in Table 1. In each case, the accuracy metrics are computed on a hold-out partition of respective datasets (using the common evaluation protocol when available).

The accuracies are comparable with the state-of-the-art on most cases. However, our computational budget is limited, and we cannot use a combination of many networks, as *e.g.,* the best methods in age estimation do.

| *Stage* | *Network* | *Accuracy* |
|---|---|---|
| *Detection* | SSD-MobileNet, α=.75 | 67.2% (AP @0.5IoU) |
| *Age* | MobileNet, α=1.0 | 4.9 years (MAE) |
| *Gender* | MobileNet, α=1.0 | 88.3% (accuracy) |
| *Expression* | MobileNet, α=1.0 | 55.9% (accuracy) |

**Table 1.** Accuracies of the different stages in our system.

### V. CONCLUSIONS

The demo has been presented several times in public locations and has shown its value in illustrating the potential of modern machine learning in an easy-to-approach use case working on many levels. At best the response of non-experts has been astonished "*how did the machine know my age?*". On the other hand, there are also important contemporary research questions addressed that provoke discussion even with the most distinguished scientists.